\documentclass[10pt,twocolumn,letterpaper]{article}

\usepackage{cvpr}
\usepackage{times}
\usepackage{epsfig}
\usepackage{graphicx}
\usepackage{amsmath}
\usepackage{amssymb}

\usepackage{enumitem}
\usepackage[skip=3pt]{caption}
\usepackage[textfont=small,aboveskip=1pt,belowskip=2pt]{subcaption}
\usepackage{tabu}
\usepackage{textcomp}
\usepackage{color}
\usepackage[table]{xcolor}
\usepackage{blindtext}
\usepackage{array}
\usepackage{multirow}
\usepackage{booktabs}
\usepackage{overpic}
\usepackage{siunitx}
\usepackage[pagebackref=true,breaklinks=true,letterpaper=true,colorlinks,bookmarks=false]{hyperref}

\newcommand{\tss}[1]{\textsuperscript{#1}}

\newcommand{\email}[1]{\href{mailto:#1}{#1}}

\usepackage[hang,flushmargin]{footmisc} 
\newcommand\blfootnote[1]{%
  \begingroup
  \renewcommand\thefootnote{}\footnote{#1}%
  \addtocounter{footnote}{-1}%
  \endgroup
}

\cvprfinalcopy 

\def\cvprPaperID{0000} 

\ifcvprfinal\fi
\begin{document}

\title{NTIRE 2020 Challenge on Real Image Denoising:\\ Dataset, Methods and Results}

\author{
Abdelrahman Abdelhamed
\and Mahmoud Afifi
\and Radu Timofte 
\and Michael S. Brown 
\and Yue Cao
\and Zhilu Zhang
\and Wangmeng Zuo
\and Xiaoling Zhang
\and Jiye Liu
\and Wendong Chen
\and Changyuan Wen 
\and Meng Liu
\and Shuailin Lv
\and Yunchao Zhang 
\and Zhihong Pan 
\and Baopu Li 
\and Teng Xi 
\and Yanwen Fan
\and Xiyu Yu
\and Gang Zhang
\and Jingtuo Liu
\and Junyu Han
\and Errui Ding
\and Songhyun Yu
\and Bumjun Park
\and Jechang Jeong
\and Shuai Liu
\and Ziyao Zong
\and Nan Nan
\and Chenghua Li
\and Zengli Yang
\and Long Bao
\and Shuangquan Wang
\and Dongwoon Bai
\and Jungwon Lee
\and Youngjung Kim
\and Kyeongha Rho
\and Changyeop Shin
\and Sungho Kim
\and Pengliang Tang
\and Yiyun Zhao
\and Yuqian Zhou
\and Yuchen Fan
\and Thomas Huang
\and Zhihao Li
\and Nisarg A. Shah
\and Wei Liu
\and Qiong Yan
\and Yuzhi Zhao
\and Marcin Mo\.{z}ejko
\and Tomasz Latkowski
\and Lukasz Treszczotko
\and Micha\l{} Szafraniuk
\and Krzysztof Trojanowski
\and Yanhong Wu
\and Pablo Navarrete Michelini
\and Fengshuo Hu
\and Yunhua Lu
\and Sujin Kim
\and Wonjin Kim
\and Jaayeon Lee
\and Jang-Hwan Choi
\and Magauiya Zhussip
\and Azamat Khassenov
\and Jong Hyun Kim
\and Hwechul Cho
\and Priya Kansal
\and Sabari Nathan
\and Zhangyu Ye
\and Xiwen Lu 
\and Yaqi Wu 
\and Jiangxin Yang 
\and Yanlong Cao 
\and Siliang Tang 
\and Yanpeng Cao 
\and Matteo Maggioni
\and Ioannis Marras
\and Thomas Tanay
\and Gregory Slabaugh
\and Youliang Yan
\and Myungjoo Kang
\and Han-Soo Choi
\and Kyungmin Song
\and Shusong Xu
\and Xiaomu Lu
\and Tingniao Wang
\and Chunxia Lei
\and Bin Liu
\and Rajat Gupta
\and Vineet Kumar 
\vspace{-1.5cm}
}

\maketitle
\begin{abstract}
This paper reviews the NTIRE 2020 challenge on real image denoising with focus on the newly introduced dataset, the proposed methods and their results. The challenge is a new version of the previous NTIRE 2019 challenge on real image denoising that was based on the SIDD benchmark. This challenge is based on a newly collected validation and testing image datasets, and hence, named SIDD+. This challenge has two tracks for quantitatively evaluating image denoising performance in (1) the Bayer-pattern rawRGB and (2) the standard RGB (sRGB) color spaces. Each track $\sim$250 registered participants. A total of 22 teams, proposing 24 methods, competed in the final phase of the challenge. The proposed methods by the participating teams represent the current state-of-the-art performance in image denoising targeting real noisy images. The newly collected SIDD+ datasets are publicly available at:  \url{https://bit.ly/siddplus\_data}.
\end{abstract}



\vspace{-9mm}
\blfootnote{A. Abdelhamed (kamel@eecs.yorku.ca, York University), M. Afifi, R. Timofte, and M.S. Brown are the NTIRE 2020 challenge organizers, while the other authors participated in the challenge. Appendix~\ref{app:teams} contains the authors' teams and affiliations.
NTIRE webpage:\\ \url{https://data.vision.ee.ethz.ch/cvl/ntire20/}.}

\vspace{9mm}
\section{Introduction}
\label{sec:introduction}

Image denoising is a fundamental and active research area (\eg,~\cite{tai2017memnet, zhang2017beyond, zhang2018ffdnet, gu2019brief}) with a long-standing history in computer vision (\eg,~\cite{kuan1985adaptive, liu2008automatic}). A primary goal of image denoising is to remove or correct for noise in an image, either for aesthetic purposes, or to help improve other downstream tasks. For many years, researchers have primarily relied on synthetic noisy image for developing and evaluating image denoisers, especially the additive white Gaussian noise (AWGN)---\eg,~\cite{BuadesCVPR05, Dabov07imagedenoising, zhang2017beyond}. Recently, more focus has been given to evaluating image denoisers on real noisy images~\cite{abdelhamed2018high, plotz2017benchmarking,abdelhamed2019ntire}.  To this end, we have proposed this challenge as a means to evaluate and benchmark image denoisers on real noisy images.

This challenge is a new version of the Smartphone Image Denoising Dataset (SIDD) benchmark~\cite{abdelhamed2018high} with a newly collected validation and testing datasets, hence named \textit{SIDD+}. The original SIDD consisted of thousands of real noisy images with estimated ground-truth, in both raw sensor data (rawRGB) and standard RGB (sRGB) color spaces. Hence, in this challenge, we provide two tracks for benchmarking image denoisers in both rawRGB and sRGB color spaces. We present more details on both tracks in the next section.

\section{The Challenge}
\label{sec:challenge}

This challenge is one of the NTIRE 2020 associated challenges on: deblurring~\cite{nah2020ntire}, nonhomogeneous dehazing~\cite{ancuti2020ntire}, perceptual extreme super-resolution~\cite{zhang2020ntire}, video quality mapping~\cite{fuoli2020ntire}, real image denoising~\cite{abdelhamed2020ntire}, real-world super-resolution~\cite{lugmayr2020ntire}, spectral reconstruction from RGB image~\cite{arad2020ntire} and demoireing~\cite{yuan2020demoireing}. 

The NTIRE 2020 Real Image Denoising Challenge is an extension of the previous NTIRE 2019 challenge~\cite{abdelhamed2019ntire}. Both challenges aimed to gauge and advance the state-of-the-art in image denoising. The focus of the challenge is on evaluating image denoisers on {\it real}, rather than synthetic, noisy images. In the following, we present some details about the new dataset used in this version of the challenge and how the challenge is designed.

\subsection{Dataset}

The SIDD dataset~\cite{abdelhamed2018high} was used for providing training images for the challenge. The SIDD dataset consists of thousands of real noisy images and their corresponding ground truth, from ten different scenes, captured repeatedly with five different smartphone cameras under different lighting conditions and ISO levels. The ISO levels ranged from 50 to 10,000. The images are provided in both rawRGB and sRGB color spaces. 

For validation and testing, we collected a new dataset of 2048 images following a similar procedure to the one used in generating the SIDD validation and testing datasets.

\subsection{Challenge Design and Tracks}

\paragraph{Tracks}

We provide two tracks to benchmark the proposed image denoisers based on the two different color spaces: the \textbf{rawRGB} and the \textbf{sRGB}. 
Images in the rawRGB format represent minimally processed images obtained directly from the camera's sensor.  These images are in a sensor-dependent 
color space where the R, G, and B values are related to the sensor's color filter array's spectral sensitivity to incoming visible light.   
Images in the sRGB format represent the camera's rawRGB image that have been processed by the in-camera image processing pipeline to map the sensor-dependent RGB colors to a device-independent color space, namely standard RGB (sRGB).  Different camera models apply their own proprietary photo-finishing routines, including several nonlinear color manipulations, to modify the rawRGB values to appear visually appealing (see~\cite{Hakki2016} for more details).  We note that the provided sRGB images are not compressed and therefore do not exhibit compression artifacts. Denoising a rawRGB would typically represent a denoising module applied within the in-camera image processing pipeline. Denoising an sRGB image would represent a denoising module applied after the in-camera color manipulation. As found in recent works~\cite{abdelhamed2019ntire,abdelhamed2018high, plotz2017benchmarking}, image denoisers tend to perform better in the rawRGB color space than in the sRGB color space.  However, rawRGB images are far less common than sRGB images which are easily saved in common formats, such as JPEG and PNG. Since the SIDD dataset contains both rawRGB and sRGB versions of the same image, we found it feasible to provide a separate track for denoising in each color space. Both tracks follow similar data preparation, evaluation, and competition timeline, as discussed next.

\paragraph{Data preparation}

The provided training data was the SIDD-Medium dataset that consists of 320 noisy images in both rawRGB and sRGB space with corresponding ground truth and metadata. Each noisy or ground truth image is a 2D array of normalized rawRGB values (mosaiced color filter array) in the range $[0, 1]$ in single-precision floating point format saved as Matlab .mat files. The metadata files contained dictionaries of Tiff tags for the rawRGB images, saved as .mat files.  

We collected a new validation and testing datasets following a similar procedure to the one used in SIDD~\cite{abdelhamed2018high}, and hence, we named the new dataset SIDD+.

The SIDD+ validation set consists of $1024$ noisy image blocks (\ie, croppings) form both rawRGB and sRGB images, each block is $256 \times 256$ pixels. The blocks are taken from $32$ images, $32$ blocks from each image ($32 \times 32 = 1024$). All image blocks are combined in a single 4D array of shape $[1024, 256, 256]$ where each consecutive 32 images belong to the same image, for example, the first 32 images belong to the first image, and so on. The blocks have the same number format as the training data. Similarly, the SIDD+ testing set consists of $1024$ noisy image blocks from a different set of images,  but following the same format as the validation set. Image metadata files were also provided for all $64$ images from which the validation and testing data were extracted. All newly created validation and testing datasets are publicly available.

We also provided the simulated camera pipeline used to render rawRGB images into sRGB for the SIDD dataset~\footnote{\url{https://github.com/AbdoKamel/simple-camera-pipeline}}. The provided pipeline offers a set of processing stages similar to an on-board camera pipeline. Such stages include: black level subtraction, active area cropping, white balance, color space transformation, and global tone mapping. 

\paragraph{Evaluation}

The evaluation is based on the comparison of the restored clean (denoised) images with the ground-truth images. For this we use the standard peak signal-to-noise ratio (PSNR) and, complementary, the structural similarity (SSIM) index~\cite{wang2004image} as often employed in the literature. Implementations are found in most of the image processing toolboxes. We report the average results over all image blocks provided.

For submitting the results, participants were asked to provide the denoised image blocks in a multidimensional array shaped in the same way as the input data (\ie, $[1024, 256, 256]$). In addition, participants were asked to provide additional information:  the algorithm's runtime per mega pixel (in seconds); whether the algorithm employs CPU or GPU at runtime; and whether extra metadata is used as inputs to the algorithm.

At the final stage of the challenge, the participants were asked to submit fact sheets to provide information about the teams and to describe their methods.

\paragraph{Timeline}

The challenge timeline was performed in two stages. The validation stage started on December 20, 2019. The final testing stage started on March 16, 2020. Each participant was allowed a maximum of 20 and 3 submissions during the validation and testing phases, respectively. The challenge ended on March 26, 2020.

\section{Challenge Results}
\label{sec:challenge_results}

\begin{figure}[h]
    \centering
    \includegraphics[width=\columnwidth]{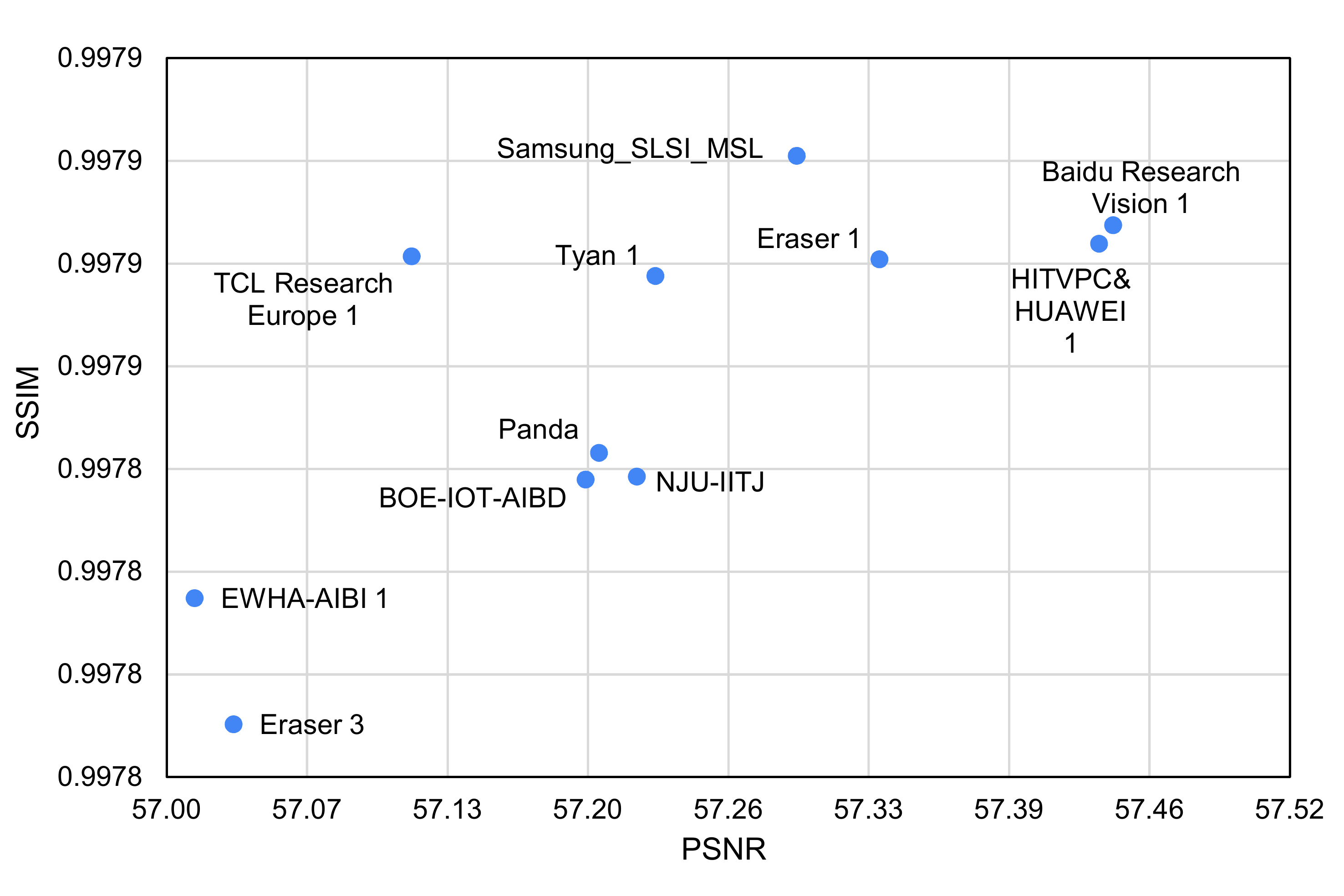}
    \caption{Combined PSNR and SSIM values of method from the rawRGB track.}
    \label{fig:results-raw}
\end{figure}

\begin{figure}[h]
    \centering
    \includegraphics[width=\columnwidth]{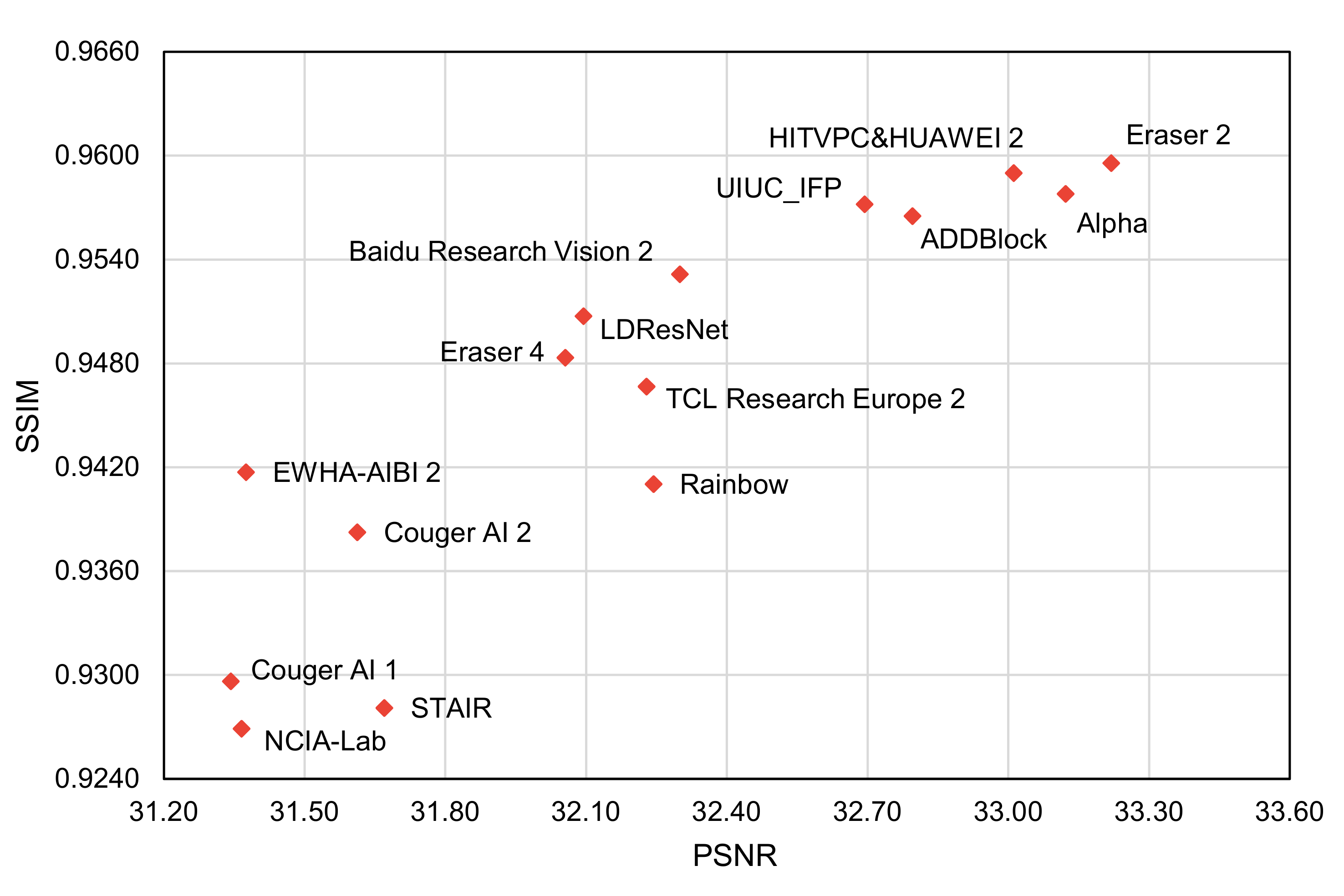}
    \caption{Combined PSNR and SSIM values of method from the sRGB track.}
    \label{fig:results-srgb}
\end{figure}

\begin{figure}
    \centering
    \includegraphics[width=\columnwidth]{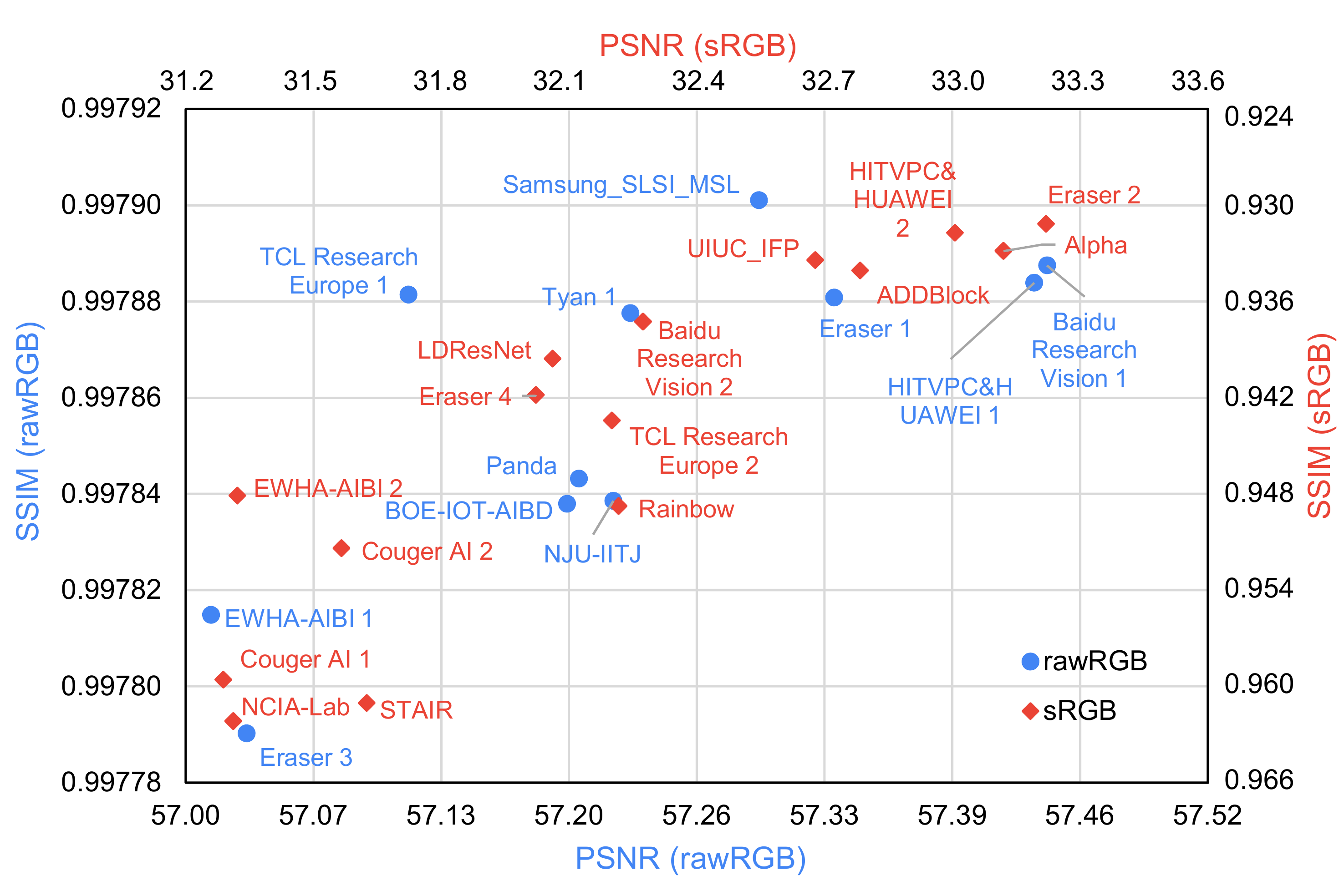}
    \caption{Combined PSNR and SSIM values of all methods from both rawRGB (in blue) and sRGB (in red) tracks. Note the different axes and scales for each track.}
    \label{fig:results-raw-srgb}
\end{figure}

From approximately $250$ registered participants in each track, $22$ teams entered in the final phase and submitted results, codes/executables, and factsheets. Tables~\ref{tab:results-raw} and~\ref{tab:results-srgb} report the final test results, in terms of peak signal-to-noise ratio (PSNR) and structural similarity (SSIM) index~\cite{wang2004image}, for the rawRGB and sRGB tracks, respectively. The tables show the method ranks based on each measure in subscripts. We present the self-reported runtimes and major details provided in the factsheets submitted by participants. Figures~\ref{fig:results-raw} and~\ref{fig:results-srgb} show a 2D visualization of PSNR and SSIM values for all methods in both rawRGB and sRGB tracks, respectively. For combined visualization, both figures are overlaid in Figure~\ref{fig:results-raw-srgb}. The methods are briefly described in section~\ref{sec:methods} and  team members are listed in Appendix~\ref{app:teams}.

\begin{table*}[t!]
\newcommand{\len}{12mm}
\newcommand{\cl}{\centering\arraybackslash}
\centering
\resizebox{\textwidth}{!}{
\begin{tabular}
{llcc >{\cl}p{15mm} >{\cl}p{20mm} >{\cl}p{20mm} >{\cl}p{27mm} >{\cl}p{20mm}}
\toprule
{Team} & {Username} & PSNR & SSIM & Runtime (s/Mpixel) & {CPU/GPU \quad (at runtime)}  & {Platform} & Ensemble & Loss\\
\midrule


Baidu Research Vision 1 & zhihongp & $57.44_{(1)}$ & $0.99789_{(2)}$ 
& $5.76$ & Tesla V100 & PaddlePaddle, PyTorch & flip/transpose ($\times 8$) & $L_1$ \\

HITVPC\&HUAWEI 1 & hitvpc\_huawei & $57.43_{(2)}$ & $0.99788_{(3)}$ 
& $?$ & GTX 1080 Ti & PyTorch & flip/rotate ($\times 8$) & $L_1$ \\ 

Eraser 1 & Songsaris & $57.33_{(3)}$ & $0.99788_{(5)}$ 
& $36.50$ & TITAN V & PyTorch & flip/rotate ($\times 8$) & $L_1$ \\ 

Samsung\_SLSI\_MSL & Samsung\_SLSI\_MSL-2 & $57.29_{(4)}$ & $0.99790_{(1)}$ 
& $50$ & Tesla V100 & PyTorch &  flip/transpose ($\times 8$), models ($\times 3$) & $L_1$ \\ 

Tyan 1 & Tyan & $57.23_{(5)}$ & $0.99788_{(6)}$ 
& $0.38$ & GTX 1080 Ti & TensorFlow & flip/rotate ($\times 8$), model snapshots ($\times 3$) & $L_1$ \\ 

NJU-IITJ & Sora & $57.22_{(6)}$ & $0.99784_{(9)}$ 
& $3.5$ & Tesla V100 & PyTorch & models ($\times 8$) & $L_1$ \\ 

Panda & panda\_ynn & $57.20_{(7)}$ & $0.99784_{(8)}$ 
& $2.72$ & GTX 2080 Ti  & TensorFlow & flip/rotate ($\times 8$), model snapshots ($\times 3$) & $L_1$ \\ 

BOE-IOT-AIBD & eastworld & $57.19_{(8)}$ & $0.99784_{(7)}$ 
& $0.61$ & Tesla P100 & TensorFlow & None & $L_1$ \\ 

TCL Research Europe 1 & tcl-research-team & $57.11_{(9)}$ & $0.99788_{(10)}$ 
& $?$ & RTX 2080 Ti & TensorFlow & flip/rotate ($\times 8$), models ($\times 3 - 5$) & $L_1$ \\ 

Eraser 3 & BumjunPark & $57.03_{(10)}$ & $0.99779_{(4)}$ 
& $0.31$ & ? & PyTorch & ? & $L_1$ \\ 

EWHA-AIBI 1 & jaayeon & $57.01_{(11)}$ & $0.99781_{(12)}$ 
& $55$ & Tesla V100 & PyTorch & flip/rotate ($\times 8$) & $L_1$ \\ 

ZJU231 & qiushizai & $56.72_{(12)}$ & $0.99752_{(11)}$ 
& $0.17$ & GTX 1080 Ti & PyTorch & self ensemble & $L_1, DCT$ \\ 

NoahDn & matteomaggioni & $56.47_{(13)}$ & $0.99749_{(14)}$ 
& $3.54$ & Tesla V100 & TensorFlow & flip/rotate ($\times 8$) & $L_1$ \\ 

Dahua\_isp & - & $56.20_{(14)}$ & $0.99749_{(13)}$ 
& $?$ & GTX 2080 & PyTorch & ? & $?$ \\ 

\bottomrule 
\end{tabular}
} 
\caption{Results and rankings of methods submitted to the rawRGB denoising track.}
 \label{tab:results-raw}
\end{table*}

\begin{table*}[t!]
\newcommand{\len}{9mm}
\newcommand{\cl}{\centering\arraybackslash}
\centering
\resizebox{\textwidth}{!}{
\begin{tabular}
{llcc >{\cl}p{15mm} >{\cl}p{20mm} >{\cl}p{20mm} >{\cl}p{25mm} >{\cl}p{22mm}}
\toprule
{Team} & {Username} & PSNR & SSIM & Runtime (s/Mpixel) & {CPU/GPU \quad (at runtime)} & {Platform}  & Ensemble & Loss\\
\midrule


Eraser 2 & Songsaris & $33.22_{(1)}$ & $0.9596_{(1)}$ 
& $103.92$ & TITAN V & PyTorch & flip/rotate/RGB shuffle ($\times 48$) & $L_1$ \\ 

Alpha & q935970314 & $33.12_{(2)}$ & $0.9578_{(3)}$ 
& $6.72$ & RTX 2080 Ti & PyTorch & flip/rotate ($\times 8$) & Charbonnier \\ 

HITVPC\&HUAWEI 2 & hitvpc\_huawei & $33.01_{(3)}$ & $0.9590_{(2)}$ 
& $?$ &  GTX 1080 Ti & PyTorch & flip/rotate ($\times 8$) & $L_1$ \\

ADDBlock & BONG & $32.80_{(4)}$ & $0.9565_{(5)}$ 
& $76.80$ & Titan XP & PyTorch & flip/rotate ($\times 8$), models ($\times 4$)& $L_1$ \\ 

UIUC\_IFP & Self-Worker & $32.69_{(5)}$ & $0.9572_{(4)}$ 
& $0.61$ & Tesla V100 ($\times 2$) & PyTorch & flip/rotate ($\times 8$), models($\times 3$) & $L_1$ \\ 

Baidu Research Vision 2 & zhihongp & $32.30_{(6)}$ & $0.9532_{(6)}$ 
& $9.28$ & Tesla V100 ($\times 8$) & PaddlePaddle, PyTorch & flip/transpose ($\times 8$) & $L_1$ \\

Rainbow & JiKun63 & $32.24_{(7)}$ & $0.9410_{(11)}$ 
& $2.41$ &  RTX 2080Ti & PyTorch & flip/rotate ($\times 8$) & $L_1$/Laplace gradient \\ 

TCL Research Europe 2 & tcl-research-team & $32.23_{(8)}$ & $0.9467_{(9)}$ 
& $?$ & RTX 2080 Ti & TensorFlow & flip/rotate ($\times 8$), models ($\times 3 - 5$) & $L_1$ \\ 

LDResNet & SJKim & $32.09_{(9)}$ & $0.9507_{(7)}$ 
& $17.85$ & GTX 1080 & PyTorch & flip/rotate ($\times 8$) & $L_1$ \\ 

Eraser 4 & BumjunPark & $32.06_{(10)}$ & $0.9484_{(8)}$ 
& $?$ & ? & PyTorch & ? & $L_1$ \\ 

STAIR & dark\_1im1ess & $31.67_{(11)}$ & $0.9281_{(14)}$ 
& $0$ & 1.86 & Titan TITAN RTX ($\times 2$) & ? & $L_1$ \\ 

Couger AI 2 & priyakansal & $31.61_{(12)}$ & $0.9383_{(12)}$ 
& $0.23$ & GTX 1080 & Keras/Tensorflow & None & MSE/SSIM \\ 

EWHA-AIBI 2 & jaayeon & $31.38_{(13)}$ & $0.9417_{(10)}$ 
& $?$ & Tesla V100 & PyTorch & flip/rotate ($\times 8$) & $L_1$ \\ 

NCIA-Lab & Han-Soo-Choi & $31.37_{(14)}$ & $0.9269_{(15)}$ 
& $2.92$ &  TITAN RTX & PyTorch & None & MS-SSIM/$L1$ \\ 

Couger AI 1 & sabarinathan & $31.34_{(15)}$ & $0.9296_{(13)}$ 
&  $0.23$  & GTX 1080 & Keras/Tensorflow & None & MSE/SSIM \\ 

Visionaries & rajatguptakgp & $19.97_{(16)}$ & $0.6791_{(16)}$
& ? & GTX 1050 Ti & PyTorch & None & MSE \\

\bottomrule 
\end{tabular}
} 
\caption{Results and rankings of methods submitted to the sRGB denoising track.}
 \label{tab:results-srgb}
\end{table*}

\paragraph{Main ideas}
All of the proposed methods are based on deep learning. Specifically, all methods employ convolutional neural networks (CNNs) based on various architectures. Most of adapted architectures are based on widely-used networks, such as U-Net~\cite{ronneberger2015unet}, ResNet~\cite{he2016deep}, and DenseNet~\cite{huang2017densely}. The main ideas included re-structuring existing networks, introducing skip connections, introducing residual connections, and using densely connected components. Other strategies have been used such as feature attention for image denoising~\cite{anwar2019real}, atrous spatial pyramid pooling (ASPP)~\cite{chen2018encoder}, and neural architectural search (NAS)~\cite{elsken2018neural}.

Most teams used $L_1$ loss as the optimization function while some teams used $L_2$ loss or adopted a mixed loss between $L_1$, $L_2$, multi-scale structural similarity (MS-SSIM)~\cite{wang2003multiscale}, and/or Laplace gradients. 

\paragraph{Top results}

The top methods achieved very close performances, in terms of PSNR and SSIM. In the rawRGB track, the top two methods are $0.01$ dB apart in terms of PSNR, whereas in the sRGB track, the top three methods have  $\sim 0.1$ dB difference in terms of PSNR,  as shown in Figures~\ref{fig:results-raw} and~\ref{fig:results-srgb}. The differences in SSIM values were similarly close.
In terms of PSNR, the main performance metric used in the challenge, the best two methods for rawRGB denoising are proposed by teams Baidu Research Vision and HITVPC\&HUAWEI, and achieved 57.44 and 57.43 dB PSNR, respectively, while the best method for sRGB denoising is proposed by team Eraser and achieved 33.22 dB PSNR. 
In terms of SSIM, as a complementary performance metric, the best method for rawRGB denoising is proposed by the team Samsung\_SLSI\_MSL and achieved a SSIM index of 0.9979, while the best SSIM index for sRGB denoising is achieved by the Eraser team. 

\paragraph{Ensembles}

To boost performance, most of the methods applied different flavors of ensemble techniques. Specifically, most teams used a self-ensemble~\cite{timofte2016seven} technique where the results from eight flipped/rotated versions of the same image are averaged together. Some teams applied additional model-ensemble techniques.

\paragraph{Conclusion}

From the analysis of the presented results, we can conclude that the proposed methods achieve state-of-the-art performance in real image denoising on the SIDD+ benchmark. The top methods proposed by the top ranking teams (\ie, HITVPC\&HUAWEI, Baidu Research Vision, Eraser, and Alpha) achieve consistent performance across both color spaces---that is, rawRGB and sRGB (see Figure~\ref{fig:results-raw-srgb}). 

\section{Methods and Teams}
\label{sec:methods}



\subsection{HITVPC\&HUAWEI} 
\begin{figure}
    \centering
    \includegraphics[width=\columnwidth]{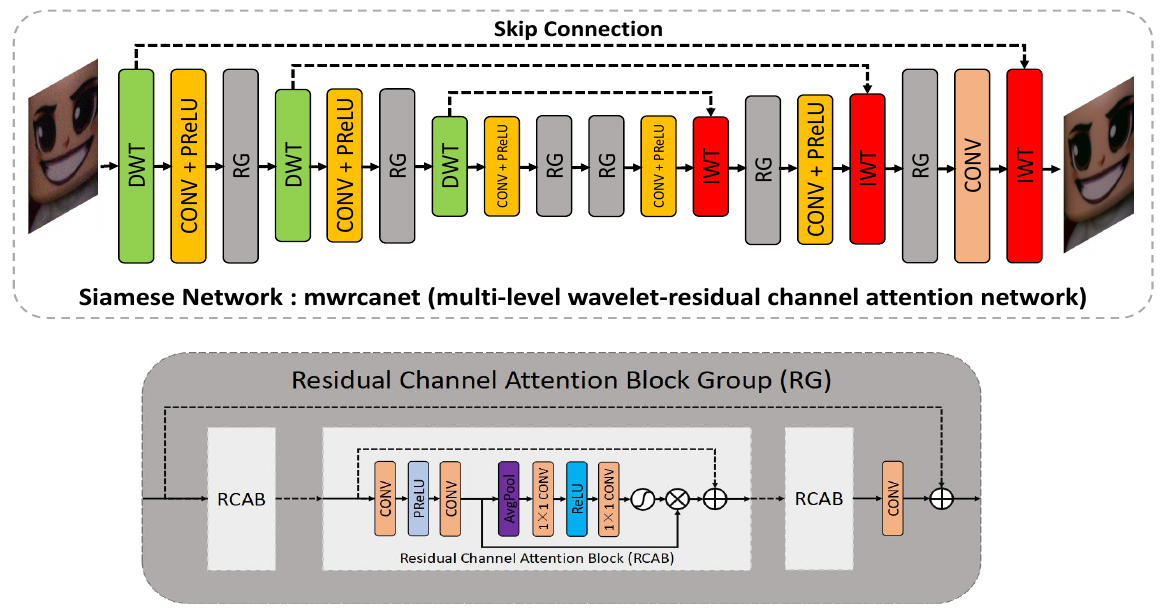}
    \caption{The network architecture proposed by the HITVPC\&HUAWEI team.}
    \label{fig:hitvpc}
\end{figure}

\paragraph{Distillating Knowledge from Original Network and Siamese Network for Real Image Denoising} The team  used
distillating knowledge and NAS (Neural Architecture Search technology) to improve the denoising performance.  The proposed network is based on both
MWCNN~\cite{liu2018multi} and ResNet~\cite{he2016deep} to propose the mwresnet (multi-level wavelet resnet).  The team used the proposed network to design the Siamese Network by use of NAS technology. Both networks can complement each other to improve denoising performance in distillating knowledge stage. Only the Siamese Network is used in  the final denoising stage. The network architecture proposed by the team is shown in Figure~\ref{fig:hitvpc}.

\subsection{Baidu Research Vision} 
\begin{figure}
    \centering
    \includegraphics[width=\columnwidth]{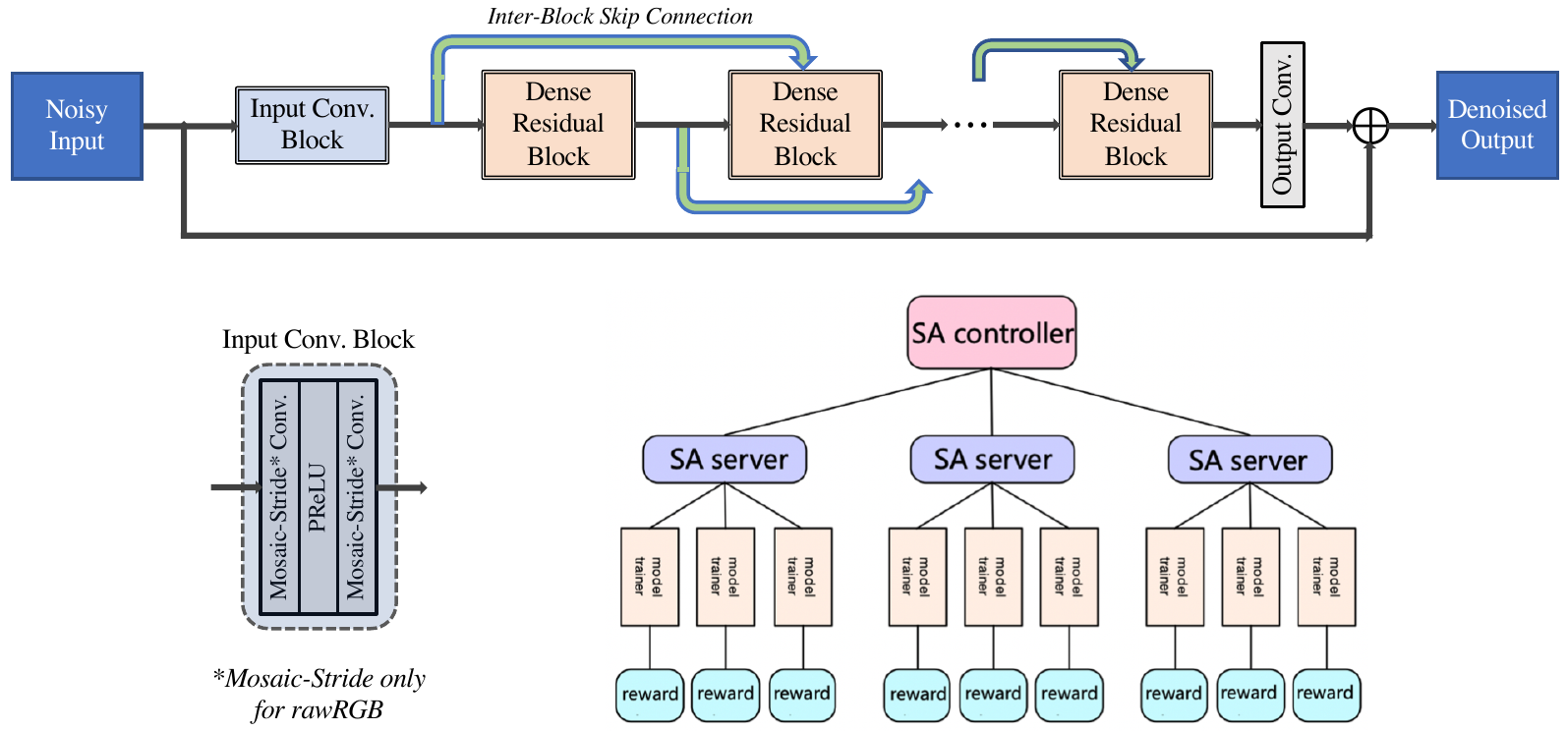}
    \caption{The architectures of the neural network and  the distributed SA-NAS scheme proposed by the Baidu Research Vision team.}
    \label{fig:baidu}
\end{figure}

\paragraph{Neural Architecture Search (NAS) based Dense Residual Network for Image Denoising}  The Baidu Research Vision team first proposed a dense residual network that includes multiple types of skip connections to learn features at different resolutions. A new NAS based scheme is further implemented in PaddlePaddle~\cite{paddlepaddle} to search for the number of dense residual blocks, the block size and the number of features, respectively. The proposed network achieves good denoising performance in the sRGB track, and the added NAS scheme achieves impressive performance in the rawRGB track. The architectures of the neural network and  the distributed SA-NAS proposed by the team are illustrated in Figure~\ref{fig:baidu}.

\subsection{Eraser} 
\begin{figure}
    \centering
    \includegraphics[width=\columnwidth]{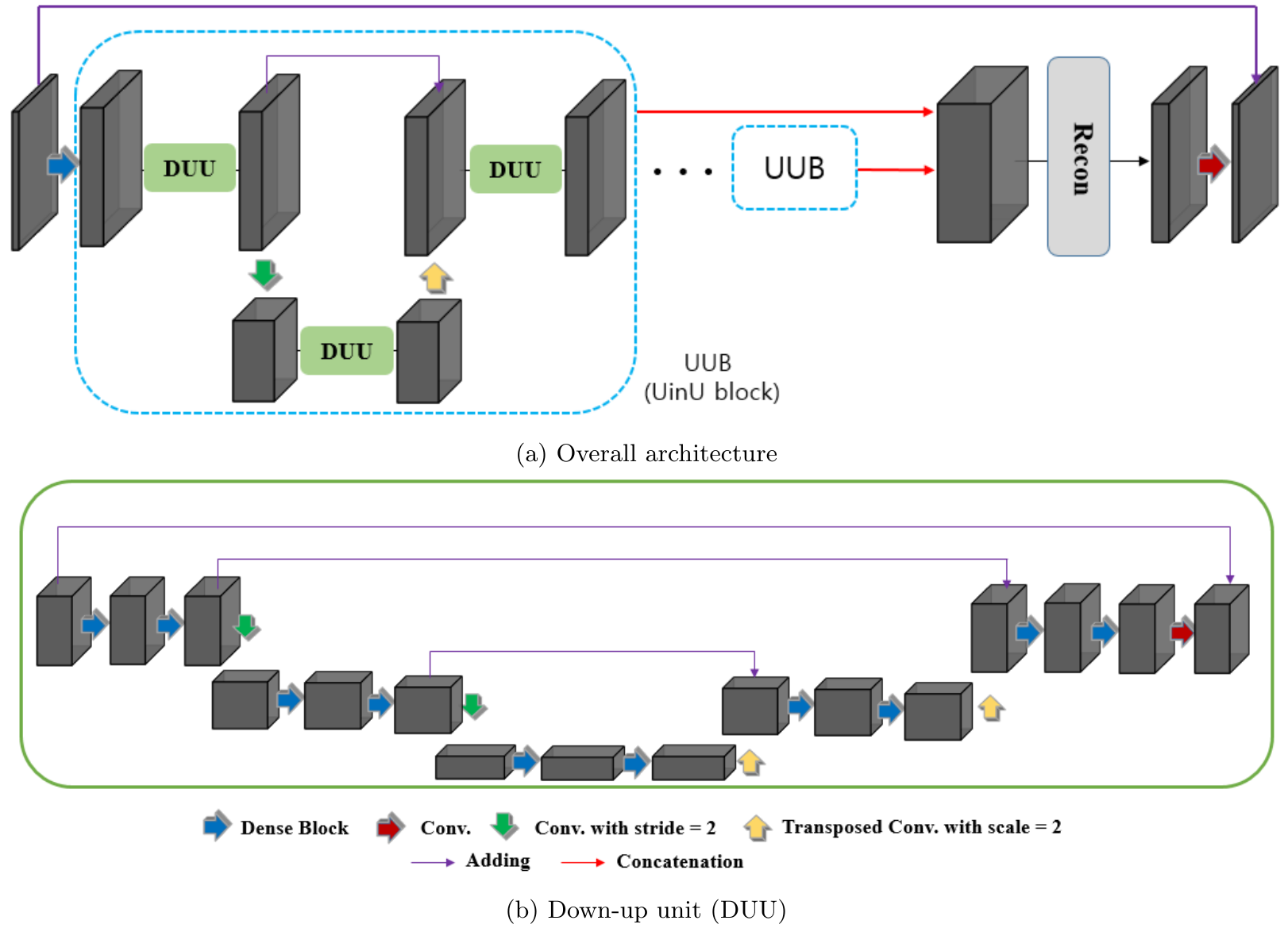}
    \caption{The UinUNet network architecture proposed by the Eraser team.}
    \label{fig:eraser-uinunet}
\end{figure}

\paragraph{Iterative U-in-U network for image denoising (UinUNet)} The team modified the down-up module and connections in the DIDN~\cite{yu2019deep}. Down-sampling and up-sampling layers are inserted between two modules to construct more hierarchical block connections. Several three-level down-up units (DUU) are included in a two-level down-up module (UUB).  The UinUNet architecture proposed by the team is shown in Figure~\ref{fig:eraser-uinunet}.

\begin{figure}
    \centering
    \includegraphics[width=\columnwidth]{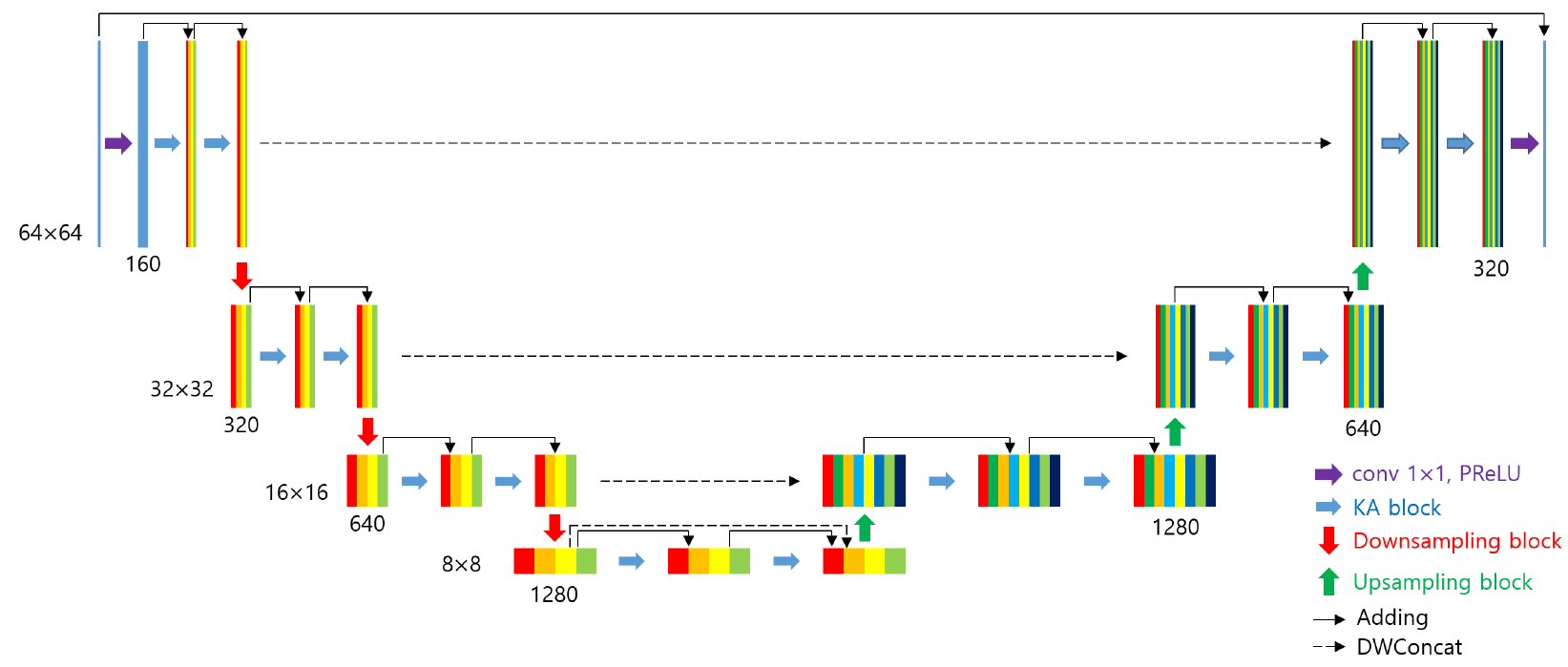}
    \caption{The KADN network architecture proposed by the Eraser team.}
    \label{fig:eraser-kadn}
\end{figure}

\paragraph{Kernel Attention CNN for Image Denoising (KADN)} is inspired by Selective Kernel Networks (SKNet)~\cite{li2019selective}, DeepLab V3~\cite{chen2018encoder}, and Densely Connected Hierarchical Network for Image Denoising (DHDN)~\cite{park2019densely}. The DCR blocks of DHDN are replaced with Kernel Attention (KA) blocks.  KA blocks
apply the concept of atrous spatial pyramid pooling (ASPP) of DeepLab V3 to apply the idea of
SKNet that dynamically selects features from different convolution kernels.
The KADN architecture proposed by the team is shown in Figure~\ref{fig:eraser-kadn}.

\subsection{Alpha} 
\paragraph{Enhanced Asymmetric Convolution Block (EACB) for image restoration tasks~\cite{Liu2020MMDM}}
Based on ACB\cite{ding2019acnet}, the team added two additional diagonal convolutions to further strengthen the kernel skeleton. For image restoration tasks, they removed the batch normalization layers and the bias parameters for better performance, and used the cosine annealing learning rate scheduler~\cite{LoshchilovSGDR} to prevent gradient explosions. They used a simple version of RCAN~\cite{zhang2018image} as the backbone. The specific modifications are: (1) Remove all channel attention (CA) modules. Too many CAs lead to an increase in training and testing time, and bring little performance improvement. (2) Remove the upsampling module to keep all the features of the same size. (3) Add a global residual to enhance the stability of the network and make the network reach higher performance in the early stages of training. 

\subsection{Samsung\_SLSI\_MSL} 
\begin{figure}
    \centering
    \includegraphics[width=\columnwidth]{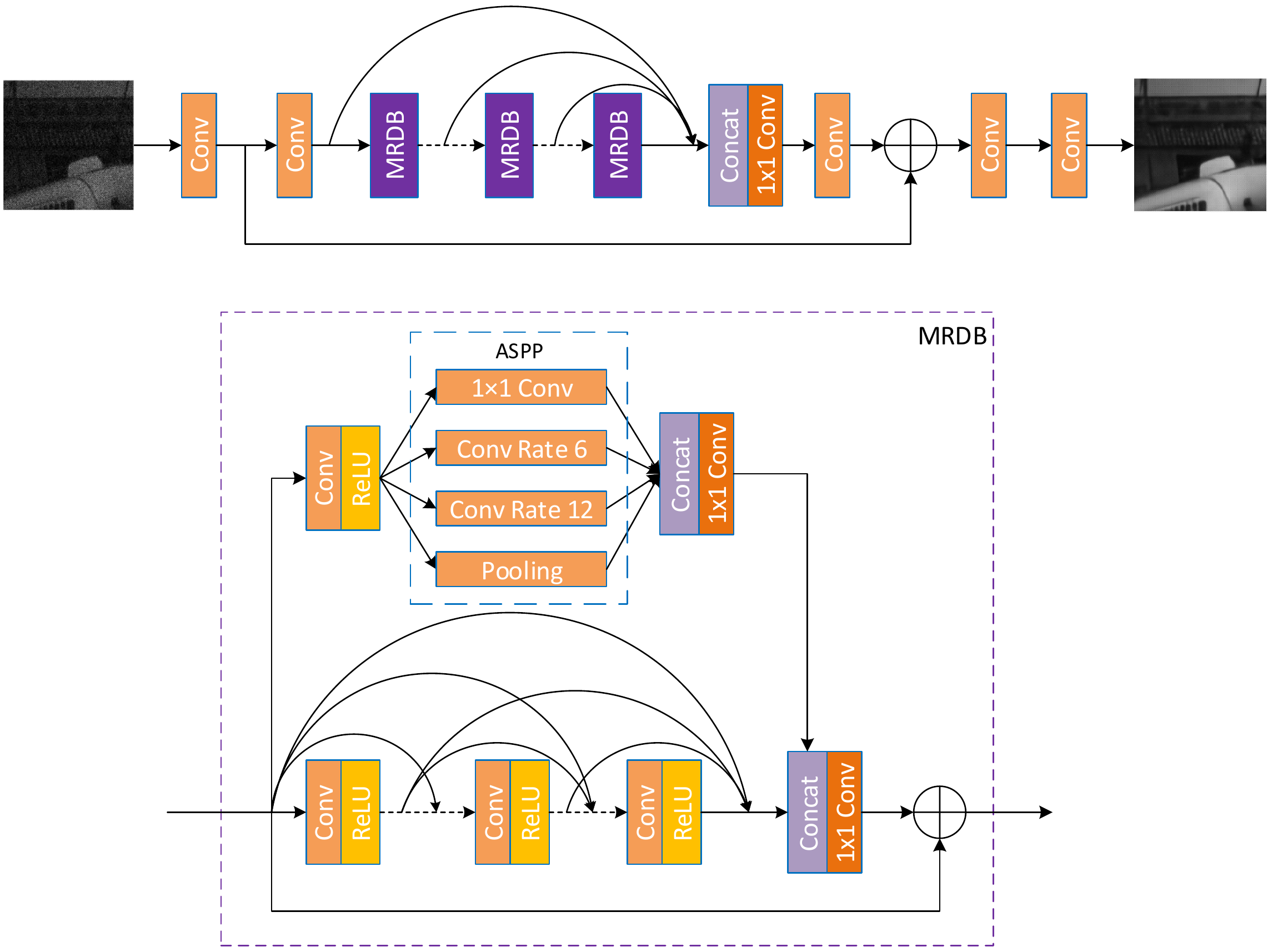}
    \caption{The MRDN architecture proposed by the Samsung\_SLSI\_MSL team.}
    \label{fig:MRDN}
\end{figure}

\paragraph{Real Image Denoising based on Multi-scale Residual Dense Block
and Cascaded U-Net with Block-connection~\cite{Bao_2020_CVPR_Workshops}} The team used three networks: residual dense network (RDN)~\cite{zhang2018residual}, multi-scale residual dense network (MRDN), and Cascaded U-Net~\cite{ronneberger2015unet} with residual dense block (RDB) connections (CU-Net). Inspired by Atrous Spatial Pyramid Pooling (ASPP)~\cite{chen2018encoder} and RDB, the team designed multi-scale RDB (MRDB) to utilize the multi-scale features within component blocks and built MRDN.  Instead of skip-connection, the team designed U-Net with block-connection (U-Net-B) to utilize an additional neural module (i.e., RDB) to connect the encoder and the decoder. They also proposed and used noise permutation for data augmentation to avoid model overfitting. The network architecture of MRDN proposed by the team is shown in Figure~\ref{fig:MRDN}, and CU-Net is detailed in~\cite{Bao_2020_CVPR_Workshops}. 

\subsection{ADDBlock} 
\begin{figure}
    \centering
    \includegraphics[width=\columnwidth]{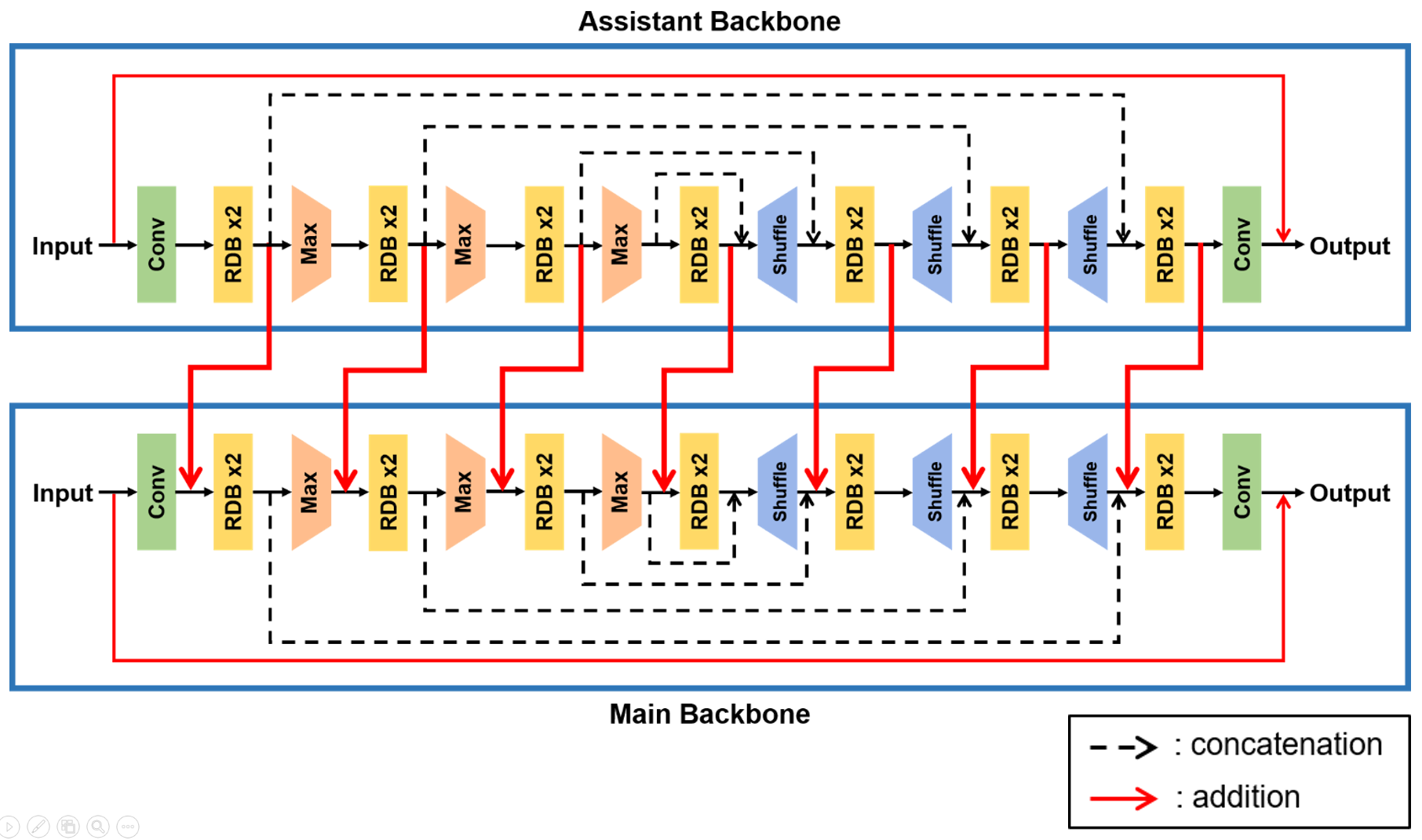}
    \caption{The PUNet architecture proposed by the ADDBlock team.}
    \label{fig:addblock}
\end{figure}

\paragraph{PolyU-Net (PUNet) for Real Image Denoising} The team utilized the idea of Composite Backbone Network (CBNet) architecture~\cite{liu2019cbnet} used for object detection. They used a U-net architecture~\cite{park2019densely} as the backbone of their PolyU-Net (PUNet). They constructed recurrent connections between backbones with only addition and without upsampling operation contrast to CBNet to prevent distortion of the original information of backbones. Additionally, contrary to CBNet, a slight performance gain was obtained by sharing weights in the backbone networks. 
The network architecture proposed by the team is shown in Figure~\ref{fig:addblock}.

\subsection{Tyan} 
\begin{figure}
    \centering
    \includegraphics[width=\columnwidth]{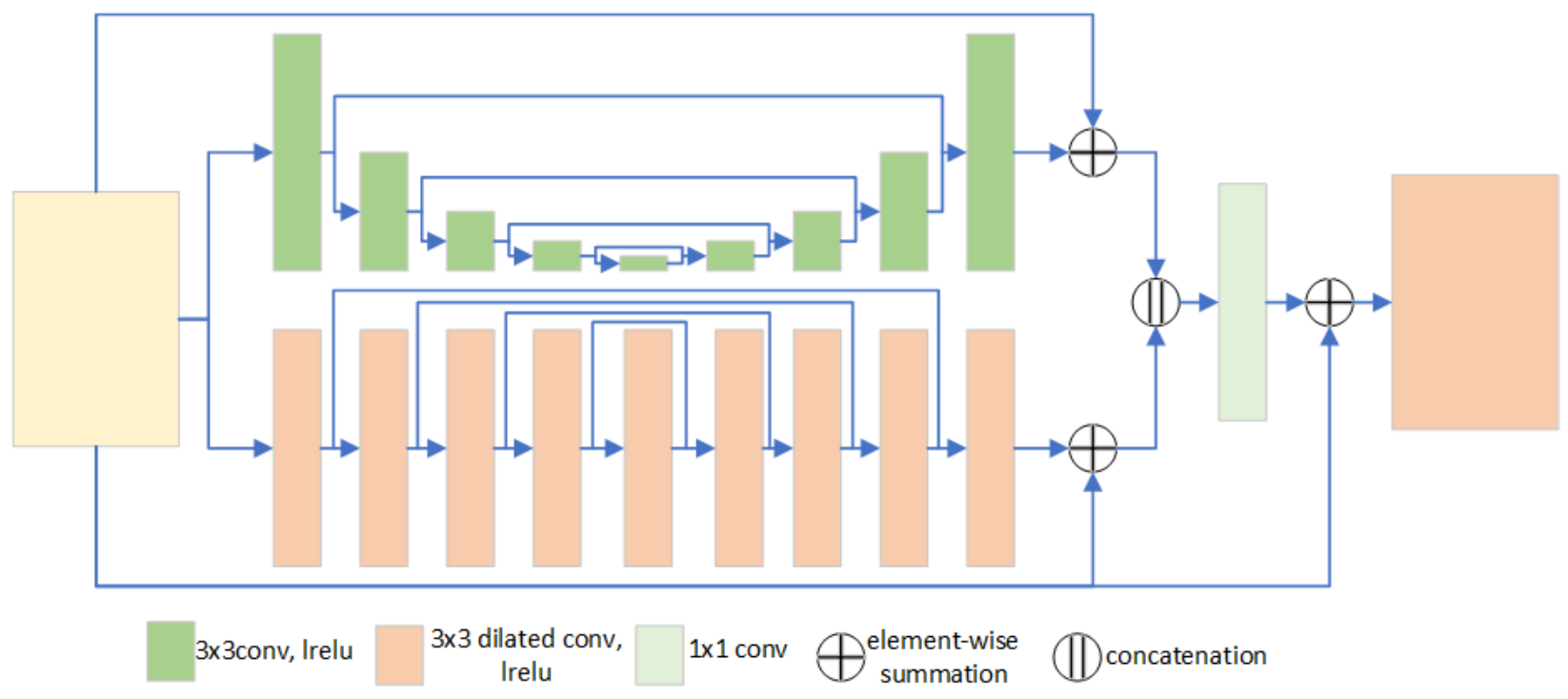}
    \caption{The Parallel U-net architecture proposed by the Tyan team.}
    \label{fig:tyan}
\end{figure}

\paragraph{Parallel U-net for Real Image Denoising} The team proposed parallel U-net for considering global and pixel-wise denoising at the same time. Two kinds of U-net were combined in a parallel way: one traditional U-net for global denoising  due to its great receptive field, and another U-net with dilated convolutions replacing the pooling operations; to preserve the feature map size. Both U-nets take the same input noisy image separately and their outputs are
concatenated and followed by a 1x1 convolution to produce the final clean image. 
The network architecture proposed by the team is shown in Figure~\ref{fig:tyan}.

\subsection{UIUC IFP} 
\paragraph{Using U-Nets as ResNet blocks for Real Image Denoising} The team concatenated multiple of the U-Net models proposed in~\cite{liu2019learning, zhou2020image}. Each U-Net model is treated as a residual block. The team used eight residual blocks in their model. Model ensemble was used to improve the performance. Specifically, the team trained ten separate models and deployed the top three models that achieves the best results on the validation set. In the testing phase, the team first applied rotating and 
flipping operations to augment each testing image. Then, a fusion operation is applied to the results obtained from the three high-performance models.

\subsection{NJU-IITJ} 
\paragraph{Learning RAW Image Denoising with Color Correction} The team adapted scaling the Bayer pattern channels based on each channel's maximum. They used Bayer unification~\cite{liu2019learning} for data augmentation and selected Deep iterative down-up CNN network (DIDN)~\cite{yu2019deep} as their base model for denoising. 

\subsection{Panda} 
\begin{figure}
    \centering
    \includegraphics[width=\columnwidth]{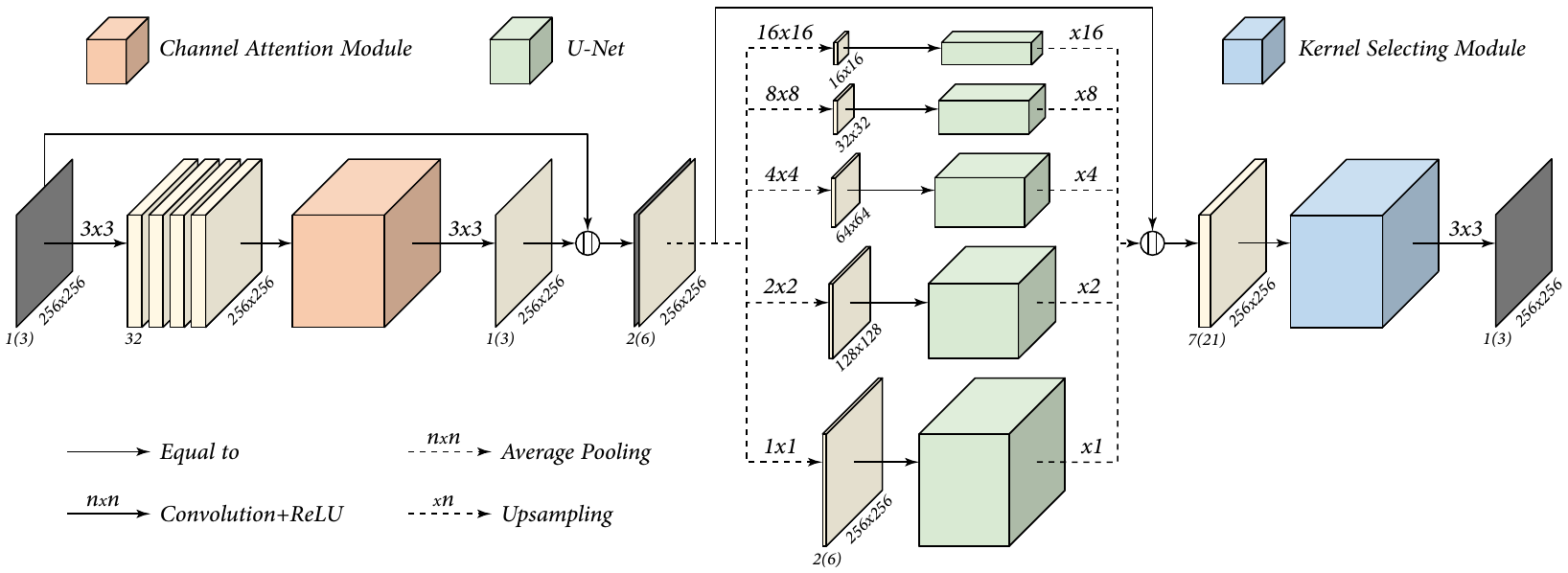}
    \caption{The PRIDNet architecture proposed by the Panda team.}
    \label{fig:panda}
\end{figure}

\paragraph{Pyramid Real Image Denoising Network} The team proposed a pyramid real image denoising network (PRIDNet), which contains three stages: (1) noise estimation stage that uses channel attention mechanism to recalibrate the channel
importance of input noise; (2) at the multi-scale denoising stage, pyramid pooling is utilized to
extract multi-scale features; and (3) the feature fusion stage adopts a kernel selecting operation to
adaptively fuse multi-scale features. 
The PRIDNet architecture proposed by the team is shown in Figure~\ref{fig:panda}.

\subsection{Rainbow} 
\begin{figure}
\centering
\includegraphics[width=\columnwidth]{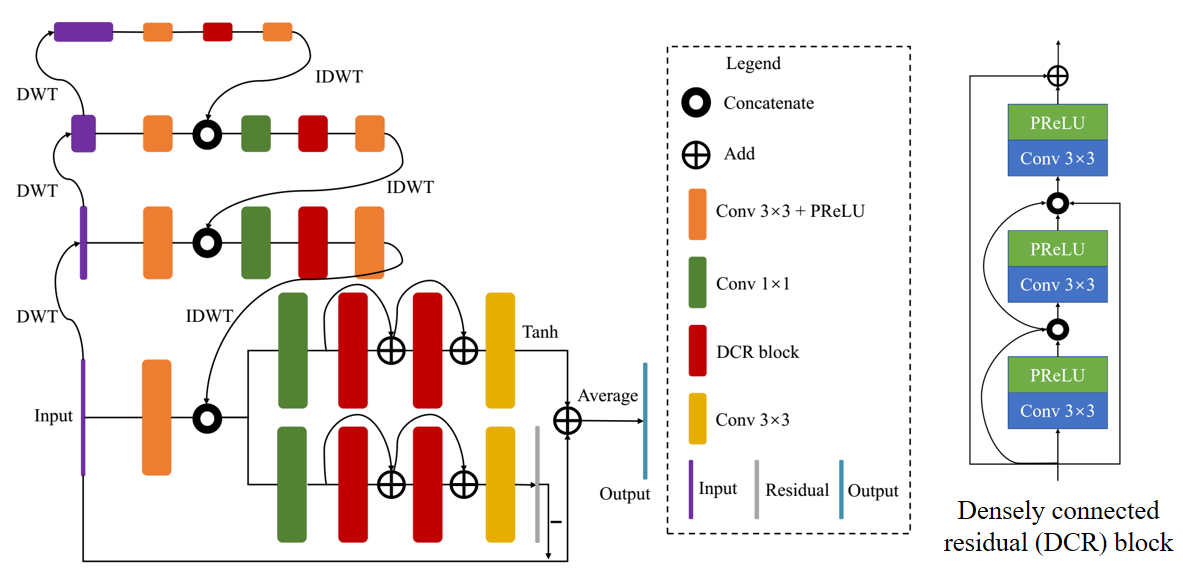}
\caption{The network architecture proposed by the Rainbow team.}
\label{fig:Rainbow}
\end{figure}

\paragraph{Densely Self-Guided Wavelet Network for Image Denoising~\cite{liu2020densely}} The team proposed a top-down self-guidance architecture for exploiting image multi-scale information. The low-resolution information is extracted and gradually propagated into the higher resolution sub-networks to guide the feature extraction processes. Instead of pixel-shuffling/unshuffling, the team used the discrete wavelet transform (DWT) and the inverse discrete wavelet transform (IDWT) for upsampling and downsampleing, respectively. The used loss was a combination between the L1 and the Laplace gradient losses.  
The network architecture proposed by the team is shown in Figure~\ref{fig:Rainbow}.

\subsection{TCL Research Europe} 
\paragraph{Neural Architecture Search for image denoising~\cite{mozejko_superkernel_2020}} The team proposed an ensemble model consisting of 3 - 5 sub-networks. Two types of sub-networks are proposed: (1) the Superkernel-based Multi Attentional Residual U-Net and (2) the Superkernel SkipInit Residual U-Net. The superkernel method used by the team is based on~\cite{stamoulis2019single}.

\subsection{BOE-IOT-AIBD} 
\begin{figure}
    \centering
    \includegraphics[width=\columnwidth]{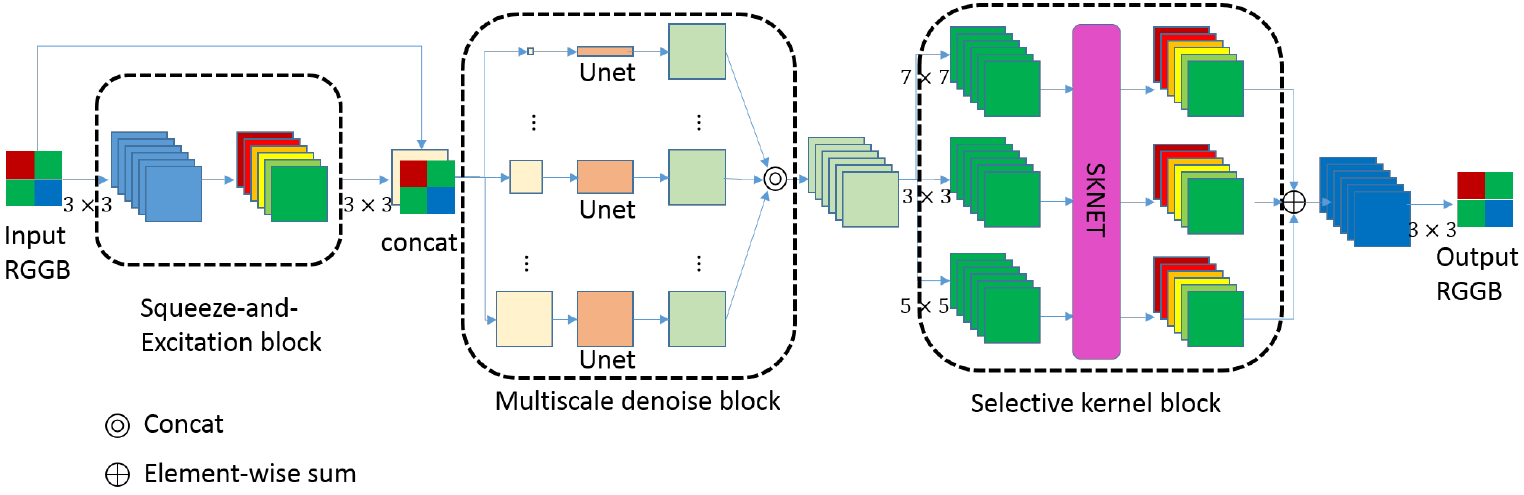}
    \caption{The network architecture proposed by the BOE-IOT-AIBD team.}
    \label{fig:BOE-IOT-AIBD}
\end{figure}

\paragraph{Raw Image Denoising with Unified Bayer Pattern and Multiscale Strategies} 
The team utilized a pyramid denoising network~\cite{zhao2019pyramid} and Bayer pattern unification techniques~\cite{liu2019learning} where all input noisy rawRGB images are unified to RGGB bayer pattern according to the metadata information. Then inputs are passed into Squeeze-and-Excitation blocks~\cite{hu2018squeeze} to extract features and assign weights to different channels. Multiscale densoising blocks and selective kernel blocks~\cite{li2019selective} were applied.
The network architecture proposed by the team is shown in Figure~\ref{fig:BOE-IOT-AIBD}.

\subsection{LDResNet} 
\begin{figure}
\centering
\includegraphics[width=\columnwidth]{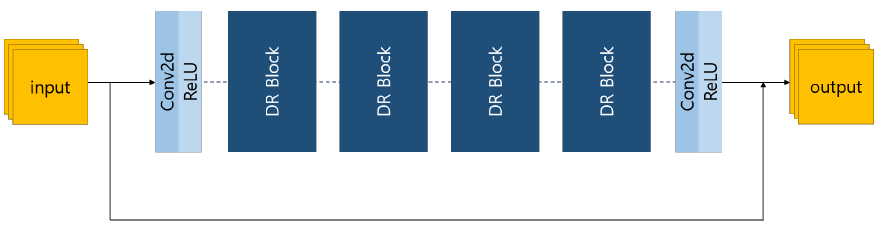}
\caption{The network architecture proposed by the LDResNet team.}
\label{fig:LDResNet}
\end{figure}

\paragraph{Mixed Dilated Residual Network for Image Denoising} The team designed a deep and wide network by piling up the dilated and residual (DR) blocks equipped with multiple dilated convolutions and skip connections. In addition to the given noisy-clean image pairs, the team utilized extra undesired-clean image pairs as a way
to add some noise on the ground truth images of the training data of the SIDD dataset. The network architecture proposed by the team is shown in Figure~\ref{fig:LDResNet}.

\subsection{EWHA-AIBI} 
\paragraph{Denoising with wavelet domain loss} The team used an enhanced deep residual network EDSR~\cite{lim2017enhanced} architecture with global residual skip and input that is decomposed with stationary wavelet transform and used loss in wavelet transform domain. To accelerate the performance of networks, channel attention~\cite{woo2018cbam} is added every fourth res block. 

\subsection{STAIR} 
\begin{figure}
\centering
\includegraphics[width=\columnwidth]{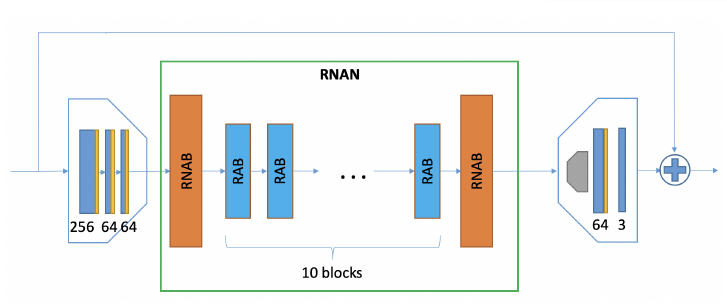}
\caption{The RNAN Down-Up scaling network (RNAN-DU) architecture proposed by the STAIR team.}
\label{fig:stair}
\end{figure}

\paragraph{Down-Up Scaling Second-Order Attention Network for Real Image Denoising} The team proposes a Down-Up scaling RNAN (RNAN-DU) method to deal with real noise that may not be statistically independent. Accordingly, the team used the residual non-local attention network (RNAN)~\cite{zhang2019residual} as a backbone of the proposed RNAN-DU method. The down-up sampling blocks are used to suppress the noise, while non-local attention modules are focused on dealing with more severe, non-uniformly distributed real noise. The network architecture proposed by the team is shown in Figure~\ref{fig:stair}.

\subsection{Couger AI} 
\paragraph{Lightweight Residual Dense net for Image Denoising}
The team proposed a U-net like model with stacked residual dense blocks along with simple convolution/convolution transpose. The input image is first processed by a coordinate convolutional layer \cite{liu2018intriguing} aiming to improve the learning of spatial features in the input image. Moreover, the team used the modified dense block to learn the global hierarchical features and then fused these features to the output of decoder in a more holistic way. 

\paragraph{Lightweight Deep Convolutional Model for Image Denoising} The team proposed also to train the network without the coordinate convolutional layer \cite{liu2018intriguing}. This modification achieves better results in the testing set compared to the original architecture. 

\subsection{ZJU231} 
\paragraph{Deep Prior Fusion Network  (DPFNet) for Real Image Denoising} The team presented DPFNet based on U-net~\cite{ronneberger2015unet}. They utilize the DPF block and the Residual block, which are both modified
versions of the standard residual block in ResNet~\cite{he2016deep}, for feature extraction and image reconstruction.
Compared with the Residual block, the DPF block introduces an extract 1 $\times$ 1 convolutional layer to
enhance the cross-channel exchange of feature maps during the feature extraction stage.

\subsection{NoahDn} 
\paragraph{Learnable Nonlocal Image Denoising} 
The team proposed a method that explicitly use the nonlocal image prior within a fully
differentiable framework. In particular, the image is processed in a block-wise fashion, and after a shallow feature extraction, self-similar blocks are extracted within a search window and then jointly denoised by exploiting their nonlocal redundancy. The final image estimate is obtained by returning each block in its original position and adaptively aggregating the overcomplete block estimates within the overlapping regions.

\subsection{NCIA-Lab} 
\paragraph{SAID: Symmetric Architecture for Image Denoising} The team proposed a two-branch bi-directional correction model. The first branch was designed to estimate positive values in the final residual layer, while the second branch was used to estimate negative values in the final residual layer. In particular, the team built their model on top of the DHDN architecture \cite{park2019densely} by adapting two models of the DHDN architecture. 

\subsection{Dahua\_isp} 
\paragraph{Dense Residual Attention Network for Image Denoising} 
The team optimized the RIDNet~\cite{anwar2019real}  with several modifications: using long dense
connection to avoid gradients vanishing; and adding depth-wise separable convolution~\cite{howard2017mobilenets} as the transition. 

\subsection{Visionaries} 
\paragraph{Image Denoising through Stacked AutoEncoders} The team used a stacked autoencoder to attempt denoising the images. While training, under each epoch, image pairs (original and noisy) were randomly shuffled and a Gaussian noise with mean and standard deviation of the difference between the original and noisy image for all 160 image pairs was added.

\section*{Acknowledgements}
We thank the NTIRE 2020 sponsors: Huawei, Oppo, Voyage81, MediaTek, DisneyResearch$\vert$Studios, and Computer Vision Lab (CVL) ETH Zurich.

\appendix
\section{Teams and Affiliations}
\label{app:teams}

\subsection*{NTIRE 2020 Team}
\noindent\textbf{Title:} NTIRE 2020 Challenge on Real Image Denoising: Dataset, Methods and Results

\noindent\textbf{Members:} \\ 
Abdelrahman Abdelhamed\tss{1} (\email{kamel@eecs.yorku.ca}), \\
Mahmoud Afifi\tss{1} (\email{mafifi@eecs.yorku.ca}), \\
Radu Timofte\tss{2} (\email{radu.timofte@vision.ee.ethz.ch}), \\
Michael S. Brown\tss{1} (\email{mbrown@eecs.yorku.ca})

\noindent\textbf{Affiliations:}\\
\tss{1} York University, Canada\\
\tss{2} ETH Zurich, Switzerland



% ...

\subsection*{HITVPC\&HUAWEI} 
\noindent\textbf{Title:} Distillating Knowledge from Original Network and Siamese Network for Real Image Denoising

\noindent\textbf{Members:}
Yue Cao\tss{1} (\email{hitvpc\_huawei@163.com}), 
Zhilu Zhang\tss{1}, 
Wangmeng Zuo\tss{1}, 
Xiaoling Zhang\tss{2}, 
Jiye Liu\tss{2}, 
Wendong Chen\tss{2}, 
Changyuan Wen\tss{2}, 
Meng Liu\tss{2}, 
Shuailin Lv\tss{2}, 
Yunchao Zhang\tss{2} 

\noindent\textbf{Affiliations:}
\tss{1} Harbin Institute of Technology, China
\tss{2} Huawei, China

\subsection*{Baidu Research Vision} 
\noindent\textbf{Title:} Neural Architecture Search (NAS) based Dense Residual Network for Image Denoising

\noindent\textbf{Members:}
Zhihong Pan\tss{1} (\email{zhihongpan@baidu.com}), 
Baopu Li\tss{1}, Teng Xi\tss{2}, Yanwen Fan\tss{2}, Xiyu Yu\tss{2}, Gang Zhang\tss{2}, Jingtuo Liu\tss{2}, Junyu Han\tss{2}, Errui Ding\tss{2}

\noindent\textbf{Affiliations:}
\tss{1} Baidu Research (USA), 
\tss{2} Department of Computer Vision Technology (VIS), Baidu Incorporation

\subsection*{Eraser} 
\noindent\textbf{Title:} Iterative U-in-U network for image denoising, Kernel Attention CNN for Image Denoising

\noindent\textbf{Members:}
Songhyun Yu (\email{3069song@naver.com}), 
Bumjun Park, Jechang Jeong

\noindent\textbf{Affiliations:}
Hanyang University, Seoul, Korea

\subsection*{Alpha} 
\noindent\textbf{Title:} Enhanced Asymmetric Convolution Block (EACB) for image restoration tasks

\noindent\textbf{Members:}
Shuai Liu \tss{1} (\email{18601200232@163.com}), 
Ziyao Zong\tss{1}, Nan Nan\tss{1}, Chenghua Li\tss{2}

\noindent\textbf{Affiliations:}
\tss{1} North China University of Technology, \\
\tss{2} Institute of Automation, Chinese Academy of Sciences

\subsection*{Samsung\_SLSI\_MSL} 
\noindent\textbf{Title:} Real Image Denoising based on Multi-scale Residual Dense Block and Cascaded U-Net with Block-connection

\noindent\textbf{Members:}
Zengli Yang (\email{zengli.y@samsung.com}), 
Long Bao, Shuangquan Wang, Dongwoon Bai, Jungwon Lee

\noindent\textbf{Affiliations:}
Samsung Semiconductor, Inc.

\subsection*{ADDBlock} 
\noindent\textbf{Title:} PolyU-Net (PUNet) for Real Image Denoising

\noindent\textbf{Members:}
Youngjung Kim (\email{read12300@add.re.kr}), 
Kyeongha Rho, Changyeop Shin, Sungho Kim

\noindent\textbf{Affiliations:}
Agency for Defense Development

\subsection*{Tyan} 
\noindent\textbf{Title:} Parallel U-Net for Real Image Denoising

\noindent\textbf{Members:}
Pengliang Tang (\email{tpl21200@outlook.com}), 
Yiyun Zhao

\noindent\textbf{Affiliations:}
Beijing University of Posts and Telecommunications

 \subsection*{UIUC IFP} 
 \noindent\textbf{Title:} Using U-Nets as ResNet blocks for Real Image Denoising

 \noindent\textbf{Members:}
 Yuqian Zhou (\email{zhouyuqian133@gmail.com}), 
 Yuchen Fan, Thomas Huang

\noindent\textbf{Affiliations:}
University of Illinois at Urbana Champaign

\subsection*{NJU-IITJ} 
\noindent\textbf{Title:} Learning RAW Image Denoising with Color Correction

\noindent\textbf{Members:}
Zhihao Li\tss{1} (\email{lizhihao6@outlook.com}), 
Nisarg A. Shah\tss{2} 

\noindent\textbf{Affiliations:}
\tss{1} Nanjing University, Nanjing, China, 
\tss{2} Indian Instititute of Technology, Jodhpur, Rajasthan, India

\subsection*{Panda} 
\noindent\textbf{Title:} Pyramid Real Image Denoising Network

\noindent\textbf{Members:}
Yiyun Zhao (\email{yiyunzhao@bupt.edu.cn}), 
Pengliang Tang

\noindent\textbf{Affiliations:}
Beijing University of Posts and Telecommunications

\subsection*{Rainbow} 
\noindent\textbf{Title:} Densely Self-guided Wavelet Network for Image Denoising

\noindent\textbf{Members:}
Wei Liu (\email{liujikun@hit.edu.cn}), 
Qiong Yan, Yuzhi Zhao

\noindent\textbf{Affiliations:}
SenseTime Research; Harbin Institute of Technology

\subsection*{TCL Research Europe} 
\noindent\textbf{Title:} Neural Architecture Search
for image denoising

\noindent\textbf{Members:}
Marcin Mo\.{z}ejko
(\email{marcin.mozejko@tcl.com}), 
Tomasz Latkowski, 
Lukasz Treszczotko, 
Micha\l{} Szafraniuk, 
Krzysztof Trojanowski

\noindent\textbf{Affiliations:}
TCL Research Europe

\subsection*{BOE-IOT-AIBD} 
\noindent\textbf{Title:} Raw Image Denoising with Unified Bayer Pattern and Multiscale Strategies

\noindent\textbf{Members:}
Yanhong Wu (\email{wuyanhong@boe.com.cn}), 

Pablo Navarrete Michelini,
Fengshuo Hu,
Yunhua Lu

\noindent\textbf{Affiliations:}
Artificial Intelligence and Big Data Research Institute, BOE

\subsection*{LDResNet} 
\noindent\textbf{Title:} Mixed Dilated Residual Network for Image Denoising

\noindent\textbf{Members:}
Sujin Kim (\email{sujin.kim@snu.ac.kr})

\noindent\textbf{Affiliations:}
Seoul National University, South Korea

\subsection*{ST Unitas AI Research (STAIR)} 
\noindent\textbf{Title:} Down-Up Scaling Second-Order Attention Network for Real Image Denoising

\noindent\textbf{Members:}
Magauiya Zhussip (\email{magauiya@stunitas.com}), 
Azamat Khassenov, Jong Hyun Kim, Hwechul Cho

\noindent\textbf{Affiliations:}
ST Unitas

\subsection*{EWHA-AIBI} 
\noindent\textbf{Title:} Denoising with wavelet domain loss

\noindent\textbf{Members:}
Wonjin Kim (\email{onejean81@gmail.com}), 
Jaayeon Lee, Jang-Hwan Choi

\noindent\textbf{Affiliations:}
Ewha Womans University

\subsection*{Couger AI} 
\noindent\textbf{Title:} Lightweight Residual Dense net for Image Denoising, Lightweight Deep Convolutional Model for Image Denoising

\noindent\textbf{Members:}
Priya Kansal (\email{priya@couger.co.jp}), 
Sabari Nathan (\email{sabari@couger.co.jp})

\noindent\textbf{Affiliations:}
Couger Inc.

\subsection*{ZJU231} 
\noindent\textbf{Title:} Deep Prior Fusion Network for Real Image Denoising

\noindent\textbf{Members:}
Zhangyu Ye\tss{1} (\email{qiushizai@zju.edu.cn}), 
Xiwen Lu\tss{2}, Yaqi Wu\tss{3}, Jiangxin Yang\tss{1}, Yanlong Cao\tss{1}, Siliang Tang\tss{1}, Yanpeng Cao\tss{1}

\noindent\textbf{Affiliations:}
\tss{1} Zhejiang University, Hangzhou, China
\tss{2} Nanjing University of Aeronautics and Astronautics, Nanjing, China
\tss{3} Harbin Institute of Technology Shenzhen, Shenzhen, China

\subsection*{NoahDn} 
\noindent\textbf{Title:} Learnable Nonlocal Image Denoising

\noindent\textbf{Members:}
Matteo Maggioni \\(\email{matteo.maggioni@huawei.com}), 
Ioannis Marras, 
Thomas Tanay, 
Gregory Slabaugh, 
Youliang Yan

\noindent\textbf{Affiliations:}
Huawei Technologies Research and Development (UK) Ltd, Noah's Ark Lab London

 \subsection*{NCIA-Lab} 
 \noindent\textbf{Title:} SAID: Symmetric Architecture for Image Denoising
\noindent\textbf{Members:}
 Myungjoo Kang (\email{mkang@snu.ac.kr}), 
 Han-Soo Choi, Sujin Kim, Kyungmin Song

\noindent\textbf{Affiliations:}
 Seoul National University

\subsection*{Dahua\_isp} 
\noindent\textbf{Title:} Dense Residual Attention Network for Image Denoising

\noindent\textbf{Members:}
Shusong Xu (\email{13821177832@163.com}), 
Xiaomu Lu, Tingniao Wang, Chunxia Lei, Bin Liu

\noindent\textbf{Affiliations:}
Dahua Technology

\subsection*{Visionaries} 
\noindent\textbf{Title:} Image Denoising through Stacked AutoEncoders

\noindent\textbf{Members:}
Rajat Gupta \\(\email{rajatgba2021@email.iimcal.ac.in}), 
Vineet Kumar

\noindent\textbf{Affiliations:}
Indian Institute of Technology Kharagpur


{\small
\bibliographystyle{ieee}
\bibliography{egbib}
}

\end{document}